\documentclass[letterpaper]{article} 
\usepackage{aaai23}  
\usepackage{times}  
\usepackage{helvet}  
\usepackage{courier}  
\usepackage[hyphens]{url}  
\usepackage{graphicx} 
\usepackage{natbib}  
\usepackage{caption} 
\usepackage{algorithm}
\usepackage{comment}
\usepackage{algorithmic}
\usepackage{color}
\usepackage{amsmath,bm}
\usepackage{tabularx}
\usepackage[colorlinks,
            linkcolor=red, 
            citecolor=green, 
            ]{hyperref}

\newcolumntype{Y}{>{\centering\arraybackslash}X}
\usepackage{multirow}
\usepackage{newfloat}
\usepackage{listings}

\DeclareCaptionStyle{ruled}{labelfont=normalfont,labelsep=colon,strut=off} 
\floatstyle{ruled}
\newfloat{listing}{tb}{lst}{}
\floatname{listing}{Listing}
\title{A Benchmark and Asymmetrical-Similarity Learning for Practical Image Copy Detection}
\author {
    Wenhao Wang\textsuperscript{\rm 1,2}\footnote{Work done during an internship at Baidu Research.},
    Yifan Sun\textsuperscript{\rm 2},
    Yi Yang\textsuperscript{\rm 3}
}
\affiliations {
    \textsuperscript{\rm 1} ReLER, University of Technology Sydney
    \textsuperscript{\rm 2} Baidu Research\\
    \textsuperscript{\rm 3} Zhejiang University\\
    wangwenhao0716@gmail.com, sunyifan01@baidu.com, yangyics@zju.edu.cn
}

\usepackage{bibentry}

\begin{document}

\maketitle

\begin{abstract}
Image copy detection (ICD) aims to determine whether a query image is an \textit{edited copy} of any image from a reference set. Currently, there are very limited public benchmarks for ICD, and all overlook a critical challenge in real-world applications, \emph{i.e.}, the distraction from hard negative queries. 
Specifically, some queries are not edited copies but are \textit{inherently similar} to some reference images. These hard negative queries are easily false recognized as edited copies, therefore significantly compromising the ICD accuracy. 
This observation motivates us to build the first ICD benchmark featuring this characteristic. Based on existing ICD datasets, this paper constructs a new dataset by additionally adding $100,000$ and $24,252$ hard negative pairs into the training and test set, respectively. Moreover, this paper further reveals a unique difficulty for solving the hard negative problem in ICD, \emph{i.e.}, there is a fundamental conflict between current metric learning and ICD. This conflict is: the metric learning adopts symmetric distance while the edited copy is an asymmetric (unidirectional) process, \emph{e.g.}, a partial crop is close to its holistic reference image and is an edited copy, while the latter cannot be the edited copy of the former (in spite the distance is equally small). This insight results in an Asymmetrical-Similarity Learning (ASL) method, which allows the similarity in two directions (the query $\leftrightarrow$ the reference image) to be different from each other. Experimental results show that ASL outperforms state-of-the-art methods by a clear margin, confirming that solving the symmetric-asymmetric conflict is critical for ICD. The NDEC dataset and code are available at \href{https://github.com/WangWenhao0716/ASL}{ASL}.
\end{abstract}
\section{Introduction \label{sec: intro}} 
\begin{figure}[t]
\centering 
\includegraphics[width=0.45\textwidth]{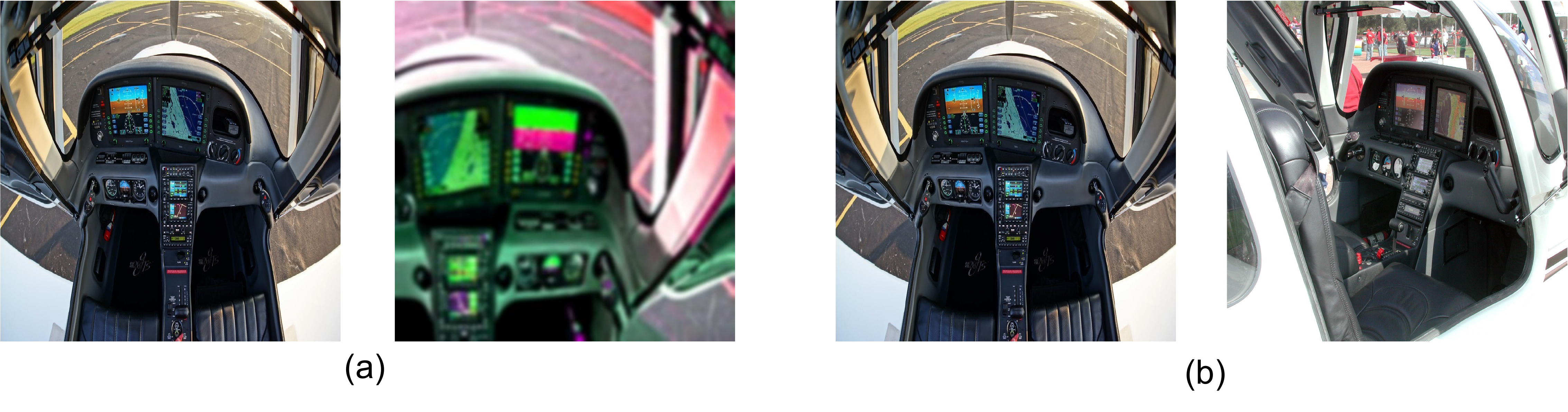}
\caption{Respective examples for edited copy and hard negative sample. In (a), the right-side image is an \textit{edited copy} of the left-side image. In (b), the right-side image is NOT the edited copy of the left-side image, in spite that they are \textit{inherently similar}. Therefore, the right-side image is a hard negative sample which easily confuses the ICD algorithm. This paper builds the first Negative Distractor for Edited Copy (NEDC) benchmark to emphasize the distraction from hard negative samples. } 
\label{vsec}
\vspace*{-4mm}
\end{figure}

Image copy detection (ICD) is to detect whether a query image is an edited copy of any image from a reference set (Fig.~\textcolor{red}{\ref{vsec} (a)}). It has extensive applications, such as preserving information legitimacy and integrity. However, current resources available for ICD research are not sufficient: the public datasets are rare and all overlook a critical challenge, \emph{i.e.}, the distraction from hard negative samples. Concretely, in realistic ICD systems, some queries are not edited copies but look inherently similar to the reference images, as shown in Fig. \textcolor{red}{\ref{vsec} (b)}. These hard negative samples are prone to being falsely recognized as edited copies and significantly compromise the ICD accuracy. Therefore, while the hard positive problem (\emph{i.e.}, some edited copies look quite dissimilar to their reference images) has already been well investigated, the hard negative problem is another critical challenge that deserves investigation in the ICD task. \par
To draw attention to the hard negative problem in ICD, this paper builds a Negative Distractors for Edited Copy (NDEC), the first benchmark featuring this characteristic from two aspects. First, NDEC adds abundant hard negative queries into the testing data, significantly increasing ICD difficulty. Among the $49,252$ queries, there are $24,252$ hard negative queries, which are quite similar to some reference images but are not edited copies. Second, the NDEC adds abundant hard negative pairs into the training data. Besides of the one million normal training images from \cite{douze20212021}, the NDEC training set contains $100,000$ additional hard negative pairs. We believe adding these hard negative pairs for training is potential to improve the ICD accuracy. In a word, \emph{the NDEC arouses the attention on hard negative distractors in ICD and meanwhile provides the prerequisite data for exploring the solutions}. 

We benchmark NDEC with state-of-the-art ICD methods and find that this new dataset is very challenging. Generally, existing ICD methods in their core rely on deep metric learning. They aim to learn a deep feature space where each edited copy and its reference image are close to each other, while the non-edited copies are far away from the reference images. Although these methods are already equipped with modern hard mining strategies, We observe that they still have difficulties to distinguish those hard negative distractors against true edited copies. This important observation indicates that in ICD, the hard negative problem is especially difficult (compared to many other metric learning tasks). \par 

\begin{figure}[t]
\centering 
\includegraphics[width=0.45\textwidth]{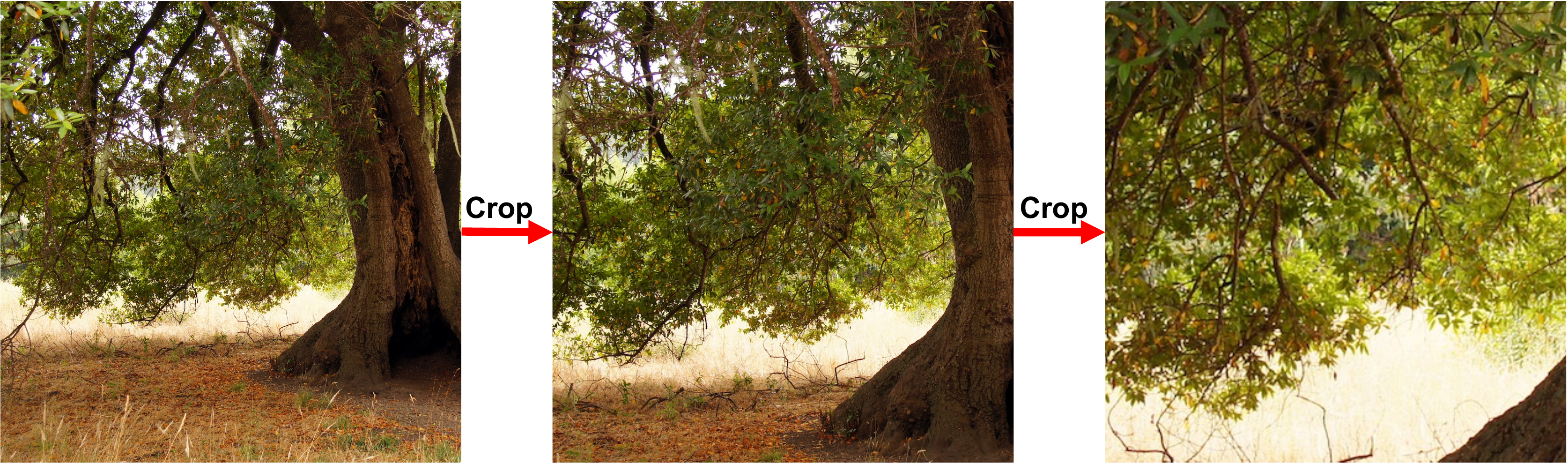}
\caption{The ``reference $\rightarrow$ edited copy'' is an unidirectional process. From left to right, the latter image is partially cropped from the former image and contains less content/information. Consequentially, the latter image is the edited copy of the former, while the former image can NOT be the edited copy of the latter. Contrary to this asymmetric process, the distance by its definition is symmetric: the distance from the former to the latter is equal to the distance from the latter to the former. }
\label{fig: unidirection}
\vspace*{-4mm}
\end{figure}

An important reason causing the above-described difficulty is: there is a fundamental conflict between current metric learning and ICD. The metric learning generally adopts symmetric distance while the edited copy is an asymmetric (unidirectional) process. For example,
a partial crop is close to its holistic reference image and is an edited copy, as shown in Fig. \textcolor{red}{\ref{fig: unidirection}}. However, in spite that the reference image is equally close to the partial crop, it cannot be the edited copy of the partial crop. More generally, the ``reference $\rightarrow$ edited copy'' process usually discards some  content/information and is thus unidirectional. Therefore, the symmetric distance is not sufficient for identifying an edited copy, especially when two images are close to each other. \par  
In response to the symmetric-asymmetric conflict, this paper proposes a novel Asymmetric-Similarity Learning (ASL) method for ICD. ASL desires the deep feature of the image containing more content/information to have larger norm, which is learnable. Correspondingly, the norm ratio between the reference and its edited copy is larger than 1, while the norm ratio between the edited copy and the reference is smaller than 1. Therefore, ASL is asymmetric and well fits the unidirectional property of edited copy. Experimental results show that ASL substantially improves ICD accuracy on NDEC. It confirms that solving the symmetric-asymmetric conflict is beneficial for realistic ICD. 

To sum up, this paper makes the following contributions:
\begin{enumerate}
 \item We contribute a new ICD dataset, \emph{i.e.}, Negative-Distractor for Edited Copy (NDEC), with emphasis on the seldom-noticed hard negative problem (while preserving the popular hard positive problem). 
 \item We benchmark NDEC with state-of-the-art methods and correspondingly reveals a fundamental conflict between the commonly-adopted symmetric distance and the asymmetric ``reference $\rightarrow$ edited copy" process. 
 \item We propose a novel Asymmetric-Similarity Learning (ASL) for ICD. ASL uses the norm ratio as an asymmetric similarity metric to distinguish edited copy against hard negative samples and substantially improves ICD. 
\end{enumerate}

\section{Related Works}
\subsection{Existing Image Copy Detection Methods}
Currently, public ICD methods are rare because organizations intend to keep the techniques as obscure as possible \cite{douze20212021}. Before deep learning era, ICD relies on hand-crafted methods, \emph{e.g.}, global descriptor \cite{kim2003content,wan2008survey}, local descriptor, Bag-of-Words models and spatial constraints \cite{zhou2010spatial,zhou2016effective}.
Deep learning has provided ICD with substantial improvement. In a recent large-scale ICD competition DISC2021 \cite{douze20212021} held by Meta AI Research, all the winners use deep learning models. Basically, they follow distance-based deep metric learning approach. For example, D$^2$LV \cite{wang2021d} and BoT \cite{wang2021bag} form each image in the training set as its own class and then use triplet loss \cite{hermans2017defense} and classification loss to train models. EfNet \cite{papadakis2021producing} uses ArcFace \cite{deng2019arcface} to gain better representations. In this paper, we find that the new NDEC dataset significantly challenges these distance-based metric learning methods with a unique conflict between the unidirectional edited copy and the symmetric distance.

\subsection{Self-supervised Learning}
So far as we know, existing ICD methods rely on self-supervised learning \cite{he2020momentum,grill2020bootstrap,chen2020simple,pizzi2022self} due to a common sense, \emph{i.e.}, the edited copies for training can be automatically generated based on unlabeled images. Specifically, the popular pipeline uses automatic transformations to generate positive training pairs and then apply deep metric learning. \par 
In contrast to this common sense, we find that automatically-generated positive pairs and self-supervised learning do not suffice to learn discriminative ICD algorithms because they do not provide hard negative pairs. Therefore, we provide $100,000$ manually-annotated hard negative pairs in addition to the unlabeled samples in NDEC and propose ASL to utilize these manually-annotated images. ASL combines both self-supervised learning (on the unlabeled data) and supervised learning (on the paired samples) and improves ICD. 

\subsection{Norm-aware Learning}
The proposed ASL learns the norm ratio between two deep features. We clarify its novelty against several works that also learns feature norm. 
Specifically, NAE \cite{chen2020norm} uses the feature's norm to separate persons and backgrounds to perform a person search task. The motivation is to relax person inter-class distances. MagFace \cite{meng2021magface} uses the feature's norm to reveal the quality of human faces. The motivation is to prevent a model from overfitting on low-quality face samples. Our motivation and approach are quite different from them: our motivation is to solve the symmetric-asymmetric conflict existing between distance-based deep metric learning and ICD; instead of norm, our approach assigns a larger than $1$ norm ratio between a reference image (the image with more content/information) and its edited copy (the image with less content/information).

\section{Datasets}
This section first briefly reviews the publicly available ICD datasets and then elaborately presents the proposed NDEC dataset. 

\subsection{Available Datasets \label{sec: available}} 
Currently, there are only two publicly available ICD datasets, \emph{i.e.}, CopyDays \cite{douze2009evaluation} and DISC21 \cite{douze20212021}. \par
The CopyDays dataset \cite{douze2009evaluation} was proposed in $2009$. It is a small-scaled dataset with only $3,157$ test samples ($157$ query images + $3,000$ reference images) and does not provide training data. Correspondingly, CopyDays does not accommodate deep learning methods. Moreover, the query images are easy to recognize, because the transformations for generating the edited copy are relatively simple, \emph{e.g.}, resizing, cropping, blurring and changing the contrast. Therefore, CopyDays does not well present the challenge in realistic ICD systems. \par
Another existing dataset DISC21 \cite{douze20212021} was proposed by Meta AI Research in NeurIPS 2021. 
Compared with CopyDays, DISC21 has three advantages. \textbf{1)} it is a large-scale dataset with one million (unlabeled) training images. \textbf{2)} it provides a challenging test set that better reflects the reality (than CopyDays does). Specifically, it has one million reference images and $50,000$ queries produced through sophisticated transformations. Many query images look quite dissimilar to their corresponding reference images and are thus hard positive samples. \textbf{3)} it mimics the realistic needle-in-haystack setting, \textit{i.e.}, the query dataset contains many distractors which are not edited copies. \par
Although DISC21 already notices the significance of distractors, its distractors are not confusing, deviating from the realistic scenario. Specifically, in DISC21, the query set does not contain hard negative samples which look very similar to the reference images. Therefore, using DISC21 for evaluation is prone to over-optimistic results. 

\begin{figure}[t]
\centering 
\includegraphics[width=0.45\textwidth]{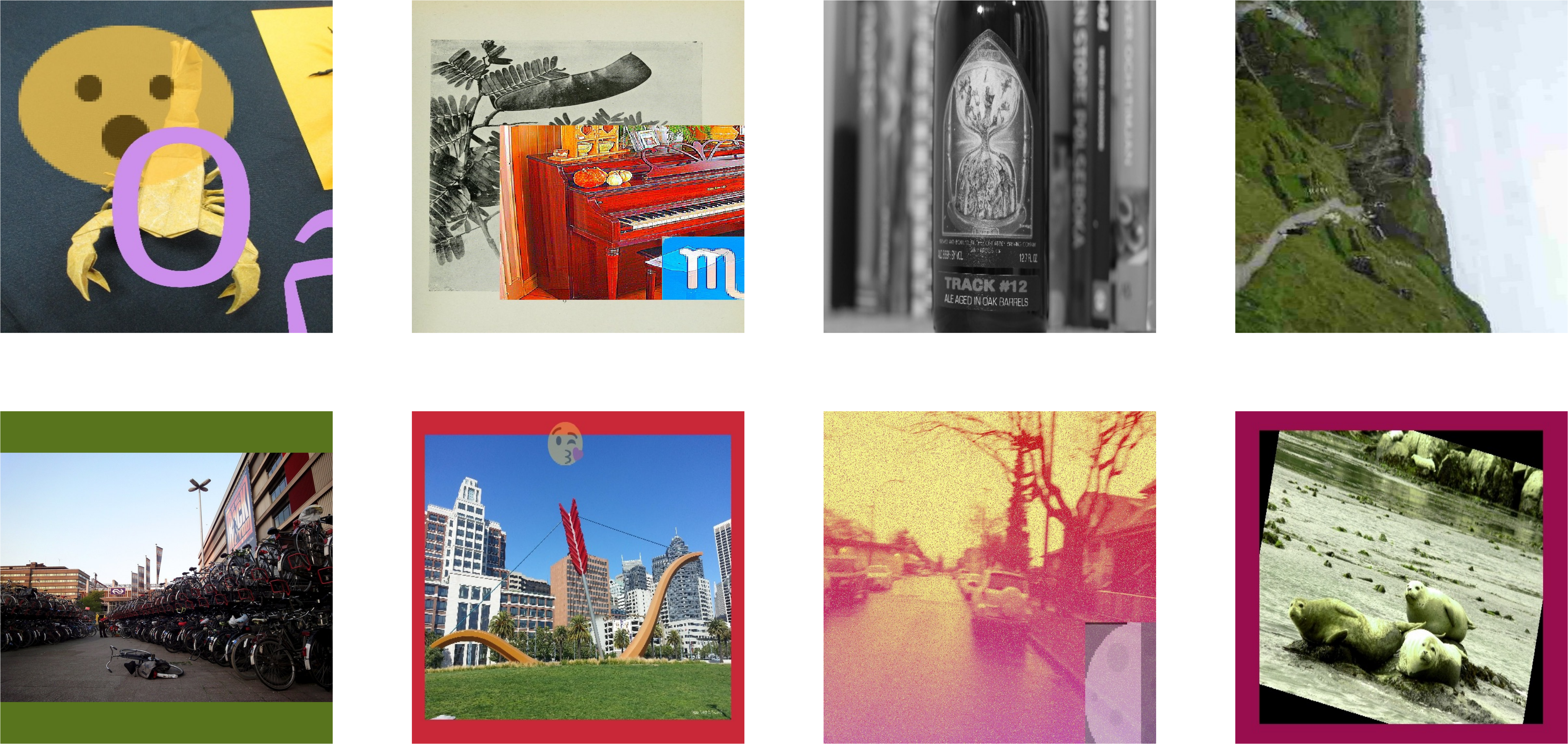}
\caption{The query images in the NDEC dataset. To simultaneously feature the hard negative and hard positive problems, we design some queries as hard negative distractors, and at the same time, there are also some edited copies with true matches in the reference dataset.} 
\label{query}
\vspace*{-4mm}
\end{figure}

\subsection{The Proposed NDEC Dataset}
To promote the research on realistic ICD, this paper proposes the NDEC (\textbf{N}egative \textbf{D}istractor for \textbf{E}dit \textbf{C}opy) dataset. \par 
The NDEC dataset is based on two sources: the DISC21 dataset \cite{douze20212021} and the OpenImage dataset \cite{OpenImages,OpenImages2}. Section \textcolor{red}{\nameref{sec: available}} already introduces DISC21. OpenImage is collected by Google and contains about nine million images annotated with image-level labels and other meta information. The NDEC does not use the annotations and merely uses OpenImage as a source for collecting samples. \par
\setlength{\parindent}{2em} \textbf{Test set.}
ICD needs to inspect whether a given query image is the edited copy of any reference image. To simulate this procedure, the test set basically consists of a query set and a reference set. In NDEC, the query and reference set contain $49,252$ and $1,000,000$ images, respectively. 
 Specifically, among the $49,252$ query images, only $5,009$ images are edited copy and have true matches in the reference set. The remaining $44,243$ query images are not edited copies but negative queries. Some of queries are shown in Fig. \textcolor{red}{\ref{query}}.\par 
Compared with the negative queries in DISC21, the negative queries in our NDEC are much more difficult: there are $24,252$ hard negative images selected from the OpenImage dataset. The selection procedure is semi-automatic: we use MultiGrain \cite{berman2019multigrain} to automatically evaluate the similarity score between OpenImage samples and the NDEC reference images. MultiGrain \cite{berman2019multigrain} returns $50,000$ candidate images that are very similar to reference images. We note that some of these $50,000$ candidate images are  true edited copies. Therefore, we filter out those edited copies from the candidates and preserve the hard negative samples through manual inspection. The manual inspection over the  $50,000$ candidate images costs $2$ annotators about $5$ days' work. \par
Another difference from DISC21 is that the NDEC prohibits training on the query and reference sets. We note that some state-of-the-art methods \cite{yokoo2021contrastive,sun20213rd} actually use $25,000$ queries from DISC21 for training and achieves extra benefits. However, in realistic ICD, using queries for training is not quite feasible. Therefore, to ensure fair evaluation on NDEC, we remove these $25,000$ queries from DISC21 when re-using its partial queries. \par
\begin{figure}[t]
\centering 
\includegraphics[width=0.45\textwidth]{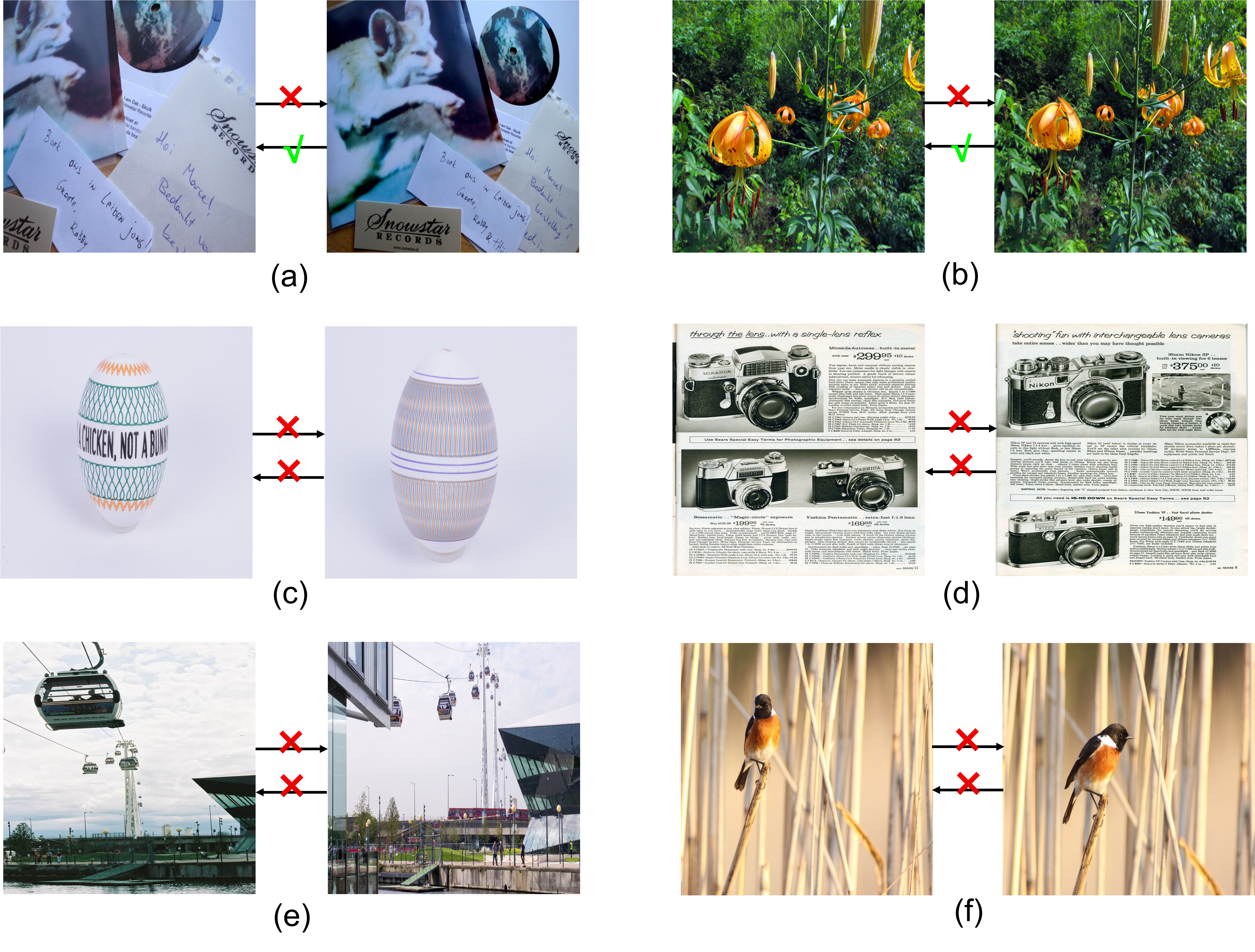}
\vspace*{-4mm}
\caption{The hard negative pairs in the training set of the NDEC dataset. In each pair, we emphasize the right-side image is not an edited copy (cropped region) of the left-side image, though the left-side image may be an edited copy of the right-side image as in (a) and (b).  In (c) and (d), these two images are visually similar. Two images from (e) are captured from different camera angles. In (f), two images are with the same location but different shooting time.} 
\label{main}
\vspace*{-4mm}
\end{figure}

 
\textbf{Training set.}
While the test set increases the ICD difficulty, the training set also provides extra hard negative pairs to promote the potential solutions. Specifically, 
the training set contains $1,100,000$ images, which are comprised of $900,000$ basic images and $100,000$ hard negative pairs (\emph{i.e.}, $100,000 \times 2$ paired images). Note that, in each pair, there is a manually labeled direction, \emph{i.e.} the latter is a hard negative image to the former, and the former may be an edited copy of the latter (See Fig. \ref{main}). The basic images are from DISC21, while the hard negative images are from OpenImage. To collect these hard negative pairs, we use the same semi-automatic procedure for preparing the hard negative queries. Concretely, we use MultiGrain \cite{berman2019multigrain} to automatically select $150,000$ candidates images (which are similar to some basic training images) and then manually filter out the edited copies. The manual filtering costs three annotators about $10$ days for initial annotation, and another three annotators $5$ days for a double-check. Fig. \textcolor{red}{\ref{main}} visualizes some hard negative pairs in the training set. 

\section{Method}
We first revisit the popular deep metric learning baseline and reveal the conflict between symmetric distance and the unidirectional reference $\rightarrow$ edited copy process. Then we propose our Asymmetrical-Similarity Learning (ASL) under an intuitive edited copy scenario (\emph{i.e.}, crop-to-copy) and generalize ASL to learn from general hard negative pairs. Finally, we emphasize our ASL with ICD baselines can handle other (basic) edited copies.
\subsection{The Limitation of Deep Metric Learning Baseline \label{sec: conflict}}
\setlength{\parindent}{2em} \hspace{1.8em}\textbf{The ICD baseline.}
Existing ICD methods in their core rely on deep metric learning.
The keynote is to learn a deep feature space where the reference image and its edited copy are close to each other, while the reference image and its non-edited copy are far away. To this end, during training, they treat a reference image and its edited copy as a positive pair, and use deep metric learning to minimize the distance within the positive pair. Concretely, the training procedure can be summarized into a two-step pipeline as below:

$\bullet$ \emph{Generating positive pairs}. Since the ICD training set does not provide ``reference-copy'' pairs but only the unpaired training samples, the training pipeline relies on multiple transformations to automatically generate the edited copies, \emph{e.g.}, cropping, blurring, and rotation. Consequentially, the auto-generated edited copy and its original image can be packed into a positive pair (or an individual class) for deep metric learning. 

$\bullet$ \emph{Deep metric learning}. Given the generated pairs or classes, deep metric learning usually uses them for pairwise training or classification training or pairwise plus classification training. The pairwise training directly optimizes the distances between samples, \emph{i.e.}, an anchor image's distance to a positive sample should be smaller than that its distance to a negative sample, through triplet loss \cite{hermans2017defense}, N-Pair loss \cite{sohn2016improved}, \emph{etc.} The classification training optimizes the distances between samples and a set of class proxies, \emph{i.e.}, a sample's distance to its target proxy should be smaller than its distance to the non-target proxies, through Softmax loss \cite{liu2016large}, CosFace \cite{wang2018cosface}, Circle loss \cite{sun2020circle}, \emph{etc.}

\textbf{The symmetric-asymmetric conflict.} A basic property of the distance is the symmetry. Let us consider a reference image $R$ and its edited copy $E$. The distance from $R$ to $E$ consistently equals the distance from $E$ to $R$, \emph{i.e.}, $D(R\rightarrow E) \equiv D(E\rightarrow R)$. Through deep metric learning, the distance from $E$ to $R$ is optimized to a small value, \emph{i.e.}, $D(E\rightarrow R) < \epsilon$ ($\epsilon$ is a threshold), indicating $E$ as an edited copy of $R$. 
Due to the symmetric property, we have $D(R\rightarrow E) < \epsilon$ as well, indicating that $R$ is an edited copy of $E$. The latter indication is incorrect, because $R$ cannot be the edited copy of $E$. As we explained in Section \textcolor{red}{\nameref{sec: intro}} and Fig.~\textcolor{red}{\ref{fig: unidirection}}, the process of reference $\rightarrow$ edited copy is unidirectional. Therefore, $R$ is close to $E$ and is a hard negative sample of $E$. Existing deep metric learning is not capable to distinguish such hard negative samples, because it 
inevitably reduces $D(R\rightarrow E)$ when it tries to reduce $D(E\rightarrow R)$. \par 
\vspace*{-1mm}
\subsection{Asymmetrical-Similarity Learning (ASL)}
\setlength{\parindent}{2em} \hspace{1.8em}\textbf{ASL under Crop-to-Copy Scenario.}
We propose Asymmetrical-Similarity Learning (ASL) to avoid the above-described symmetric-asymmetric conflict. For intuitive understanding, we first focus on the popular crop-to-copy scenario\footnote{CopyDays \cite{douze2009evaluation} points out most practical copyright violations occurs with cropping.}, where the edited copy is generated through cropping (Fig.~\textcolor{red}{\ref{fig: unidirection}}). Another reason for we specifically focus on the crop-to-copy scenario is that it can be learned through self-supervision and thus achieves accuracy gains without any annotation cost. \par 

Given two deep features $\bm{x}_i$ and $\bm{x}_j$ from image $l_i$ and $l_j$, respectively, ASL defines the ratio between their norm value as: 

\begin{equation}
    R(\bm{x}_i\rightarrow \bm{x}_j)=\left| \bm{x}_i\right|/\left| \bm{x}_j\right|
\end{equation}
\begin{figure*}[t]
\centering 
\includegraphics[width=0.75\textwidth]{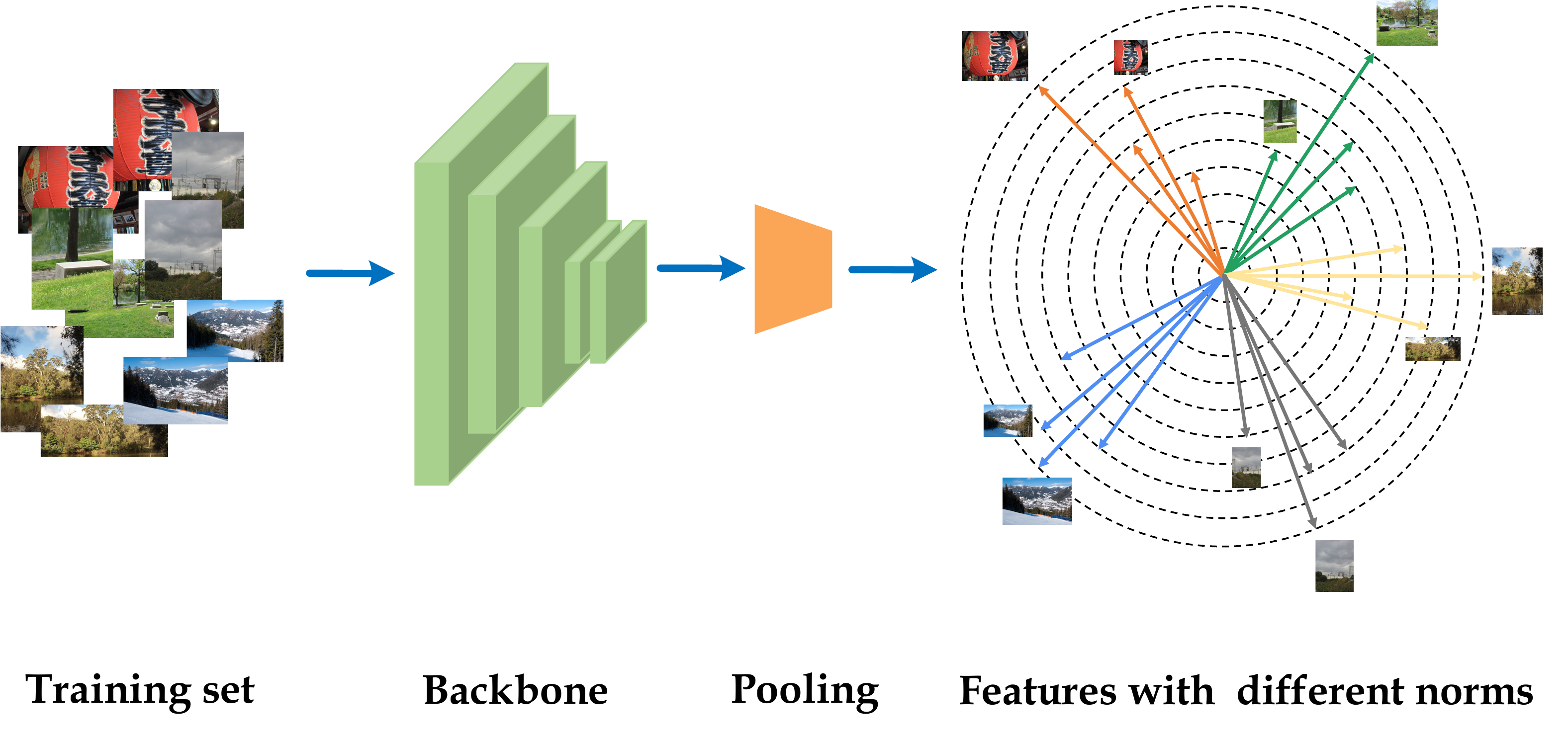}
\caption{The illustration of Asymmetrical-Similarity Learning (ASL). In ASL, distance-based metric learning loss pulls the features of inherently similar images close to each other. Meanwhile, the norm ratio based loss makes images with more content/information have a larger norm. 
For better visualization, on the right side, larger image sizes represent images with more content/information (though when training, all the images are resized to the same size).} 
\label{norm}
\vspace*{-4mm}
\end{figure*}
Obviously, the norm ratio is asymmetric, because $R(\bm{x}_i\rightarrow \bm{x}_j) \neq R(\bm{x}_j\rightarrow \bm{x}_i)$.
Given this asymmetric property, ASL uses norm ratio as the metric to identify crop-to-copy. Specifically, if $l_j$ is a cropped copy of $l_i$ (\emph{i.e.}, $l_i$ is the reference image of $l_j$), ASL desires $R(\bm{x}_i\rightarrow \bm{x}_j) > 1$. To this end, ASL generates abundant cropped copies and learns their norm ratios in a self-supervised manner, as illustrated in Fig. \textcolor{red}{\ref{norm}}. Given a reference image, ASL crops a series of sub-images with decreasing sizes. During training, ASL selects a larger crop $l_i$ and a smaller crop $l_j$ as the pseudo reference image and the edited copy, respectively. \par 

The training objective is 
\begin{equation}
\mathcal{L} =\exp{\large(1-R(\bm{x}_i\rightarrow \bm{x}_j)\large)}+\lambda \cdot \mathcal{L}_{mtr},
\label{eq: loss}
\end{equation}
where $\mathcal{L}_{mtr}$ is a popular distance-based metric learning loss and $\lambda$ is a balance parameter. $\mathcal{L}_{mtr}$ can be implemented as CosFace, where the angles between different classes are learned to be larger than a margin. During training, decreasing the loss value $\mathcal{L}$ increases $R(\bm{x}_i\rightarrow \bm{x}_j)$, therefore making the norm value of reference image $l_i$ larger than that of the edited copy $l_j$. Therefore, ASL has two simultaneous learning effects, \emph{i.e.}, it pulls inherently similar images close to each other and makes the image with more content/information have a larger feature norm, as illustrated in Fig.\textcolor{red}{~\ref{norm}}. \par 

During testing, given a query image $l_Q$ and a reference image $l_E$, we first calculate their distance to identify whether they are close to each other. If the distance between them is small, we further compare their norm ratio $R(\bm{x}_Q\rightarrow \bm{x}_E)$. If $R(\bm{x}_Q\rightarrow \bm{x}_E)>1$, it indicates that the query is actually larger than the reference image and thus is unlikely to be an edited copy. \par
\setlength{\parindent}{0em} \hspace{1.8em}\textbf{ASL Using Hard Negative Pairs.}
The ASL in before section focuses on the crop-to-edit copy scenario in a self-supervised training manner. Given the manually annotated hard negative training pairs in NDEC, we generalize ASL to learn from more scenarios. We recall that the hard negative pairs in NDEC training set are annotated in single directions, \emph{i.e.}, within each negative pair, the latter image $l_L$ is similar to, but not the edited copy of the former image $l_F$, as illustrated in Fig. ~\textcolor{red}{\ref{main}}. 

The principle of learning from these hard negative pairs is: ASL desires $R(\bm{x}_L \rightarrow \bm{x}_F) > 1$ because $l_L$ is not the edited copy of $l_F$ (the first item in Eq.~\textcolor{red}{\ref{eq: loss}}), while considering $l_L$ is positive to $l_F$ in $\mathcal{L}_{mtr}$ (the second item in Eq.~\textcolor{red}{\ref{eq: loss}}) because these two images are inherently similar, as detailed below:

$\bullet$ \textbf{The norm ratio item in ASL treats $l_L$ as negative to $l_F$.} Specifically, for the norm ratio item in Eq.~\textcolor{red}{\ref{eq: loss}}, ASL considers that $l_L$ contains no less content/information than $l_F$ because $l_L$ is not the edited copy of $l_F$. Therefore, ASL optimizes the norm ratio $R(\bm{x}_L \rightarrow \bm{x}_F)$ to be larger than $1$.

$\bullet$ \textbf{The distance-based metric learning item treats $l_L$ as positive to $l_F$}. Specifically, $\mathcal{L}_{mtr}$ in ASL actually pulls $l_L$ and $l_F$ close to each other. This is because $l_L$ and $l_F$ are manually identified as being inherently similar to each other (in spite that there is no edited copy). If $\mathcal{L}_{mtr}$ treats them as negative pair and pushes them far apart, it will make the true positive queries hard to recognize, therefore compromising ICD (\emph{w.r.t.} the recall, in particular).

We note that although the above two items seem to stand against each other (one treats $l_L$ as a negative to $l_F$ while the other treats $l_L$ as a positive to $l_F$), they actually brings joint benefits for ICD.  
We validate this design through ablation study in Table\textcolor{red}{~\ref{abla_1}}. This confirms that ASL effectively alleviates the symmetric-asymmetric conflict in ICD task.  \par 
\setlength{\parindent}{0em} \hspace{1.8em}\textbf{Combining ASL with the baseline.}
We note that ASL by itself only focuses on distinguishing the hard negative samples. Therefore, it is important to combine ASL with the ICD baseline, so as to recognize those basic edited copy patterns, \emph{e.g.} color-jittered and padded images. Consequently, the baselines recognize the basic edited copies and ASL further filters out the hard negative ones. \par 

\section{Experiments}
\subsection{Protocols and Baselines}
\hspace{1.8em}\textbf{Evaluation protocols.}
We adopt two protocols, \textit{i.e.} $\mu AP$ and precision. Following DISC21 \cite{douze20212021}, $\mu AP$ reflects the objectives of ICD. Besides $\mu AP$, we also use precision, defined as precision=TP/(TP+FP) where ``TP'' and ``FP'' denote the number of true positive and false positive matches in the top $N$ pairs, respectively. Specifically, the $N$ is set as $5,009$ (the number of queries with true matches), which causes that precision=$1$ ideally.\par
\hspace{1.8em}\textbf{Baselines.} We use two baselines, \emph{i.e.}, a strong baseline used in D$^2$LV \cite{wang2021d} and a simple (and relatively weak) baseline used in BoT \cite{wang2021bag}. The difference between these two baselines is that the former one uses additional model ensemble. The strong baseline allows ASL to compete against state-of-the-art methods, while the simple baseline is more convenient and clearer for ablation studies. To be computationally efficient, we replace the $8192$-dim features \cite{wang2021d,wang2021bag} with commonly-used $2048$-dim features. Moreover,  instead of combining cross-entropy loss and triplet loss \cite{hermans2017defense}, we use a single CosFace \cite{wang2018cosface}. On DISC21, our implementation achieves consistent results ($90.13\%$ and $72.42\%$ $\mu AP$) as the baselines in D$^2$LV ($90.10\%$ $\mu AP$) \cite{wang2021d} and BoT ($72.73\%$ $\mu AP$) \cite{wang2021bag}, respectively. \par


\vspace*{-2mm}
\subsection{The Challenge from NDEC}

\setlength{\parindent}{2em} \hspace{1.8em}\textbf{Increasing hard negative queries decreases the accuracy.}
Based on the simple baseline, we analyze how the distraction from hard negative queries impacts the ICD accuracy. We recall that the test set of NDEC
\begin{figure}[t]
\centering
 \includegraphics[width=0.35\textwidth]{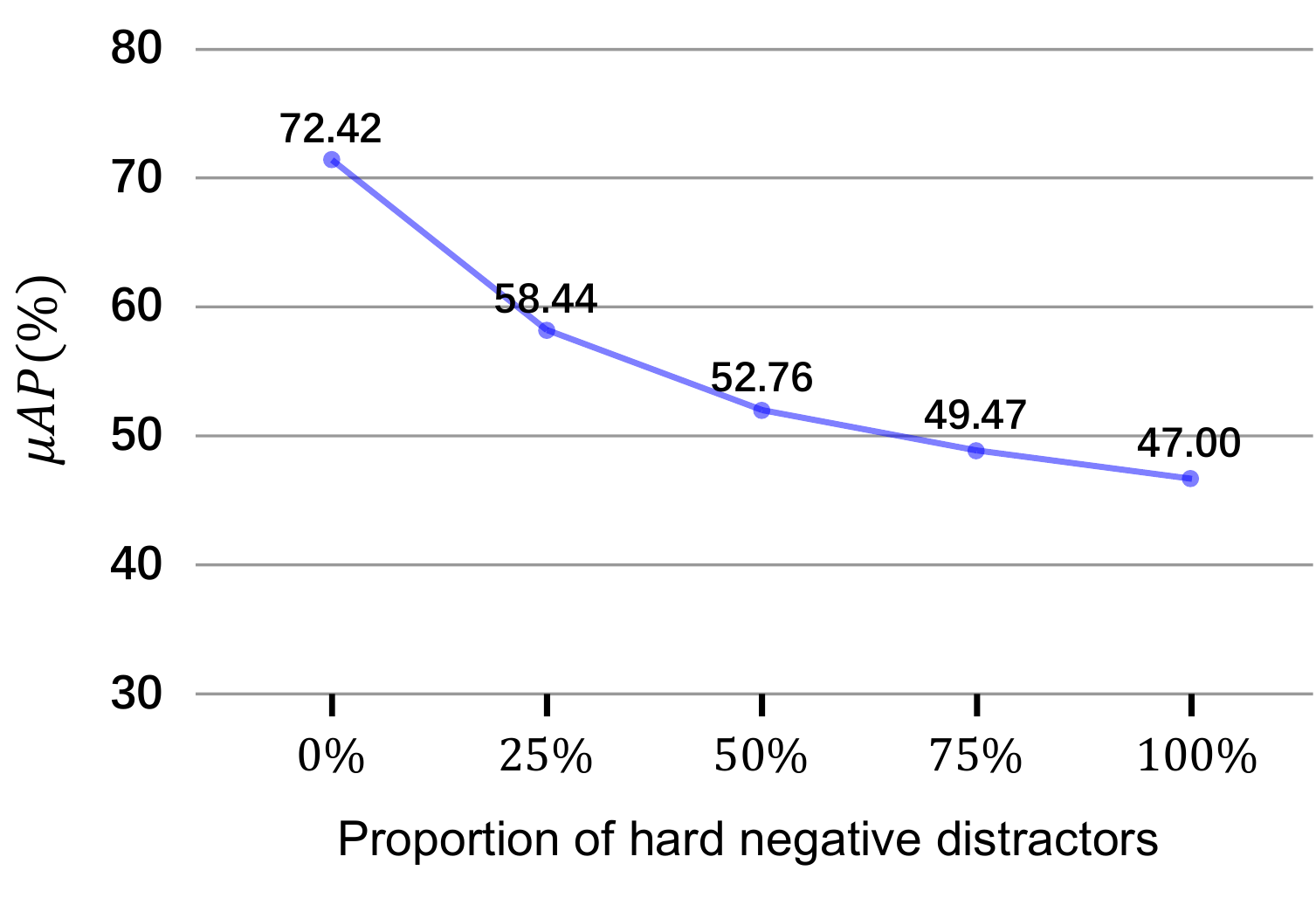}
  \vspace*{-4mm}
 \caption{The performance change of different proportions of hard negative distractors.}
 \label{perf}
 \vspace*{-6mm}
\end{figure}
contains $24,252$ hard negative queries from OpenImage and $25,000$ ($5,009$ positive 
and $19,991$ relatively easy negative) queries from  DISC21 dataset. We maintain the latter $25,000$ queries unchanged in all experiments and gradually increase the hard negative queries. 
The corresponding $\mu AP$ scores of the baseline are
summarized in Fig. \textcolor{red}{\ref{perf}}, from which we draw two observations. First, we clearly observe that as the hard negative queries increase from $0$ to $24,252$ (100\%), the $\mu AP$ scores gradually decrease from $72.42\%$ to $47.00\%$ ($-25.42\%$). It indicates that the hard negative queries are very difficult to distinguish and significantly compromise the ICD accuracy. Second, when the volume of hard negative queries approaches 100\%, the decreasing speed becomes very slow (\emph{e.g.}, $-2.47\%$ $\mu AP$ drop for the last 25\% increase of hard negative queries), indicating that the $24,252$ hard negative queries in NDEC are roughly sufficient for presenting the challenge.

\textbf{NDEC significantly challenges the state-of-the-art methods.}
We further benchmark NDEC with state-of-the-art methods \cite{wang2021d,sun20213rd,yokoo2021contrastive,papadakis2021producing,wang2021bag} (top performing methods in DISC2021 \cite{douze20212021}) in Fig. \textcolor{red}{\ref{drop}}. It clearly shows that all the methods undergo a substantial accuracy decrease when the evaluation dataset is changed from DISC21 to our NDEC. Specifically, the $\mu AP$ drops are ranged from $23.13\%$ (CNNCL) to $39.09\%$ (EsViTp), indicating that NDEC is much more challenging than DISC21. Given that all the other testing conditions on DISC21 and NDEC (except for the hard negative distraction) are roughly the same, we infer the hard negative distraction in NDEC, which reflects the realistic challenge, is the major reason for this accuracy decrease. 

\vspace*{-2mm}
\subsection{The Effectiveness of ASL}
\begin{figure}[t]
\centering 
\includegraphics[width=0.45\textwidth]{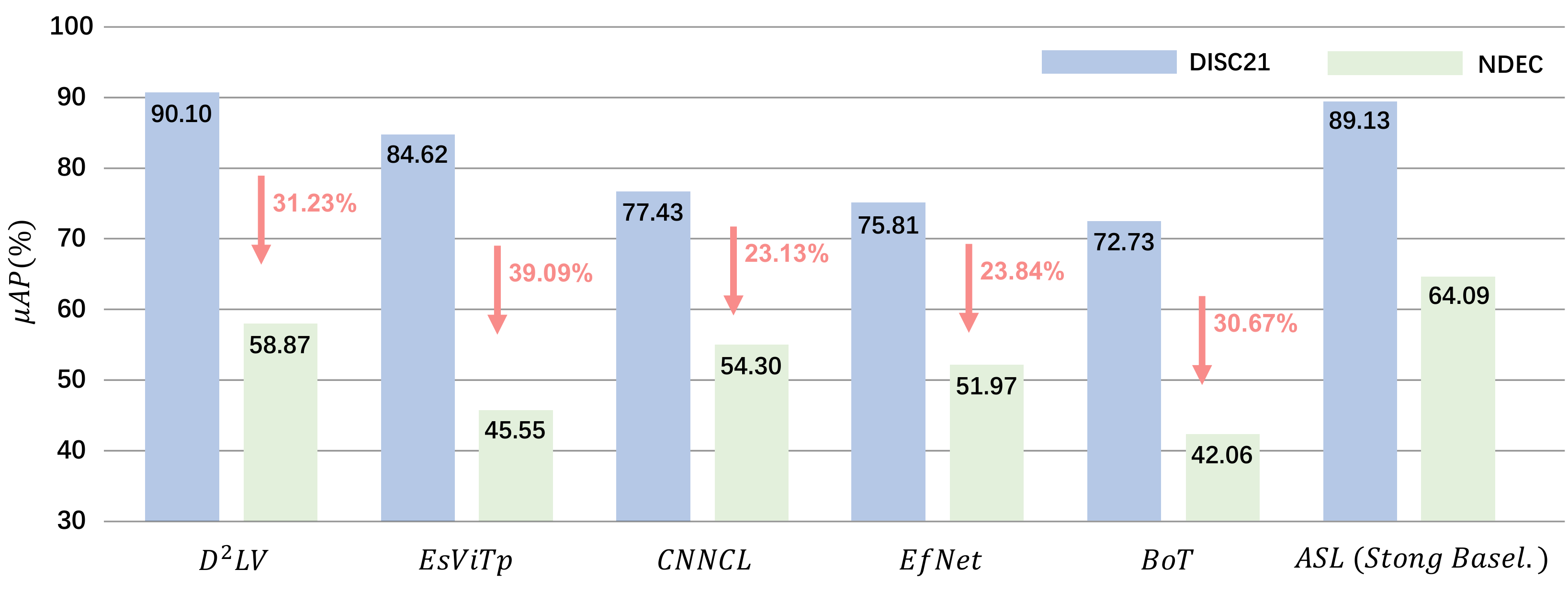}
\vspace*{-2mm} 
\caption{Evaluation on DISC21 and our NDEC. All the methods undergo a substantial accuracy decrease when the evaluation dataset is changed from DISC21 to NDEC.}
\label{drop}
\vspace*{-6mm}
\end{figure}

\hspace{1.8em} \textbf{ASL achieves new state of the art.}
Fig. \textcolor{red}{\ref{drop}} further compares the proposed ASL against state-of-the-art methods, from which we draw two observations. First, on the challenging NDEC, ASL (with strong baseline) surpasses all the competing methods by a large margin. For example, ASL is higher than D$^2$LV, the strongest competitor by $+5.22\%$  $\mu AP$. It indicates that ASL has the strongest capacity for solving the hard negative problem, which is the characteristic of NDEC as well as an important challenge in realistic ICD.\par
Second, we observe that on DISC21, ASL (with strong baseline) is not the best ($-0.97\%$ lower than D$^2$LV). In other words, although ASL has significant superiority on NDEC, this superiority does not benefits its performance on DISC21. This is reasonable because DISC21 has no hard negative queries, and D$^2$LV already well distinguishes those relatively easy negative distractors. It is another evidence that DISC21 does not well reflect the hard negative challenge in realistic ICD, and NDEC is valuable. \par 

\begin{table*}[t]
\caption{ASL brings general improvement over various baselines. In the first group, the baselines are employed in an off-the-shelf manner.} 
\label{SOO}
\vspace*{-2mm}
\small
  \begin{tabularx}{\hsize}{Y|Y|Y|Y|Y}
    \hline
Method&$\mu AP$ $\uparrow$&True Positive $\uparrow$&False Positive $\downarrow$&Precision $\uparrow$\\ 
 \hline
 EsViTp &$45.55\%$&$2,346$&$2,663$&$46.84\%$\\
 EsViTp+ASL&$48.31\%$&$1,265$&$501$&$71.63\%$\\
 \hline
 CNNCL &$54.30\%$&$2,620$&$2,389$&$52.31\%$\\
 CNNCL+ASL&$56.95\%$&$2,110$&$545$&$79.47\%$\\
 \hline
 EfNet &$51.97\%$&$2,579$&$2,430$&$51.49\%$\\
 EfNet+ASL&$53.81\%$&$1,997$&$562$&$78.04\%$\\
 \hline
 BoT &$42.06\%$&$2,075$&$2,934$&$41.43\%$\\
 BoT+ASL&$45.07\%$&$1,620$&$838$&$65.91\%$\\
 \hline
 D$^2$LV &$58.87\%$&$2,836$&$2,173$&$56.62\%$\\
 D$^2$LV+ASL&$61.28\%$&$2,227$&$583$&$79.25\%$\\
 \hline
 \hline
 Simple Basel.&$47.00\%$&$2,269$&$2,740$&$45.30\%$\\
 Simple + ASL &$49.30\%$&$1,829$&$648$&$73.84\%$\\
 \hline
 Strong Basel. &$61.03\%$&$2,968$&$2,041$&$59.25\%$\\
 Strong + ASL &$64.09\%$&$2,331$&$567$&$80.43\%$\\
 \hline
  \end{tabularx}
  \vspace*{-4mm}
  \label{abla}
  \\
\end{table*}

\begin{table*}[t]
\caption{The ablation studies based on our simple baseline.} 
\vspace*{-2mm}
\small
  \begin{tabularx}{\hsize}{Y|Y|Y|Y|Y}
    \hline
Method&$\mu AP$ $\uparrow$&True Positive $\uparrow$&False Positive $\downarrow$&Precision $\uparrow$\\ \hline
Simple Basel.&$47.00\%$&$2,269$&$2,740$&$45.30\%$\\ \hline
ASL-Crop&$49.10\%$&$2,098$&$1,233$&$62.98\%$\\
ASL-Negative&$48.14\%$&$1,877$&$845$&$68.96\%$\\
ASL-Positive&$48.17\%$&$1,932$&$794$&$70.87\%$\\
Triplet&$45.37\%$&$1,774$&$1,200$&$59.65\%$\\
 \hline
ASL&$\textbf{49.30\%}$&$1,829$&$648$&$\textbf{73.84\%}$\\ 
 \hline
  \end{tabularx}
  \vspace*{-5mm}
  \label{abla_1}
  \\
  \vspace*{-4mm}
\end{table*}

\textbf{ASL consistently improves various baselines.} 
Besides of our strong and simple baselines, we use more state-of-the-art methods as additional baselines to show that ASL brings general improvement, as illustrated in Table \textcolor{red}{\ref{SOO}}. Please note that for these additional baselines, we do not re-implement them and then integrate them into ASL. Instead, we use their models in an off-the-shelf manner and use an independently-trained ASL as a post processing. Therefore, ``BoT + ASL'' and ``D$^2$LV + ASL'' achieve different (inferior) results compared to our ``Simple + ASL'' and ``Strong + ASL'', respectively, in spite that the underlying baseline methods are the same. 

Table \textcolor{red}{\ref{SOO}} shows that ASL consistently improves most the indicators (\emph{i.e.}, $\mu AP$, false positive and precision) over all the baselines. It indicates that ASL is effective, even if we simply combine it with an off-the-shelf baseline. We also note that for most baselines, ASL marginally compromises the true positive. It is because when ASL uses the predicted norm ratio to filter the ``inherently similar'' negative queries, it inevitably false recognizes some true edited copies as negative queries. This observation should draw two-fold attention. On the one hand, given the significant improvement on ``false positive'' and precision, we think the slight degradation on true positive is worthy and the overall benefit is valuable. On the other hand, the hard negative problem in NDEC is still far from being solved and calls for more efforts from the research community. 

\vspace*{-2mm}
\subsection{Ablation Studies}
We investigate three aspects of ASL, \emph{i.e.}, ASL under crop-to-copy scenario, ASL using hard negative pairs and the superiority of norm ratio against triplet loss in Table \textcolor{red}{\ref{abla_1}}. Specifically, for the ASL using hard negative pairs, there is an important design, \emph{i.e.}, given a hard negative training pair, the distance-based metric learning item in Eq. \textcolor{red}{\ref{eq: loss}} treats it as a positive pair. We compare this design (``ASL-Positive'') against its counterpart (``ASL-Negative''), in which the metric learning item treats the negative pair as it is. From Table \textcolor{red}{\ref{abla_1}}, we draw four observations as below:

First, comparing ``ASL-Crop'' against ``Simple Basel.'', we observe ASL under crop-to-copy scenario already brings significant improvement. Specifically, it reduces the false positive matches from $2,740$ to $1,233$ and increases the precision from $45.30\%$ to $62.98\%$. 

Second, comparing ``ASL-Negative'' against ``ASL-Positive'', we find that when learning from additional hard negative samples, ``ASL-Positive'' treating these samples as positive pair is better. It is because these samples, though containing no edited copies, are inherently similar. Therefore, ASL \textbf{1)} treats them as positive pairs for distance-based metric learning to make them close to each other in the deep feature space and \textbf{2)} leaves the function of distinguishing them from true edited copies with the norm ratio item. 

Third, replacing the norm ratio item with a  canonical triplet loss is inferior. Although using a triplet loss brings some improvement \emph{w.r.t.} the false positive and precision over the baseline, it is much worse than ASL with norm ratio. It validates that the symmetric-asymmetric conflict confuses distance-based metric learning and using norm ratio is critical for ASL to alleviate this conflict. 

Fourth, comparing ``ASL'' (ASL combining crop and hard negative pairs) against other ASL editions, we observe another round of substantial improvement, \emph{e.g.}, $+10.86\%$ precision over ``ASL-Crop'' and $+ 2.97\%$ precision over ``ASL-Positive''. It indicates that both the crop-to-copy learning and the hard negative learning are beneficial and are complementary to each other.

\vspace*{-2mm}

\section{Conclusion}

This paper introduces the hard negative problem into the Image Copy Detection task and makes three contributions. First, we contribute the NDEC dataset, which highlights the hard negative distractor challenge. NDEC adds abundant hard negative queries to draw the attention to this critical challenge and meanwhile provides hard negative training pairs to promote exploring the solutions. Second, we reveal a fundamental conflict between the distance-based metric and the edited copy process, which makes the hard negative problems very difficult. Third, we propose a novel Asymmetrical-Similarity Learning method to alleviate the above conflict. ASL learns an asymmetric similarity metric based on norm ratio and substantially improves ICD. We also notice that ASL is still far from solving the hard negative challenge and call for more research efforts from the ICD community.  
\bibliography{egbib}
\end{document}